\documentclass{article}
\usepackage[utf8]{inputenc}

\usepackage{ijcai20}

\usepackage{times}

\usepackage{soul}
\usepackage{url}
\usepackage[hidelinks]{hyperref}
\usepackage[utf8]{inputenc}
\usepackage[small]{caption}
\usepackage{graphicx}

\usepackage{amsmath}
\usepackage{amsthm}
\usepackage{amsfonts}
\usepackage{amssymb}
\usepackage{mathtools}

\usepackage{booktabs}
\usepackage{algorithm}
\usepackage{algorithmic}
\graphicspath{{../figures/}{./figures}{.}{../}{..}}
\usepackage[labelformat=simple]{subcaption}

\title{How do Offline Measures for Exploration in Reinforcement Learning behave?} \author{
Jakob~J.~Hollenstein$^1$\footnote{Contact Author}\and
Sayantan~Auddy$^1$\and
Matteo~Saveriano$^{1}$\and
Erwan~Renaudo$^{1}$\And
Justus~Piater$^1$\\
\affiliations
$^1$Department of Computer Science, University of Innsbruck, Innsbruck, Austria\\
\emails
\{name.surname\}@uibk.ac.at
}

\date{May 2020}

\usepackage{biblatex}
\usepackage[inline]{enumitem}
\usepackage{graphicx}
\usepackage{amsmath}
\usepackage{amssymb}
\usepackage{url}
\usepackage{soul}
\usepackage{verbatim}

\usepackage{xcolor}

\definecolor{orchid}{rgb}{0.85, 0.44, 0.84}
\definecolor{palecopper}{rgb}{0.85, 0.54, 0.4}
\definecolor{sapgreen}{rgb}{0.31, 0.49, 0.16}
\definecolor{amber(sae/ece)}{rgb}{1.0, 0.49, 0.0}
\definecolor{mediumpersianblue}{rgb}{0.0, 0.4, 0.65}
\renewcommand{\refeq}[1]{(\ref{#1})}
\newcommand{\xurel}{\operatorname{X}_{\mathcal{U}\text{rel}}}
\newcommand{\DKL}{D_{\textrm{KL}}}
\newcommand{\xbin}{X_\text{bin}}
\newcommand{\xbbm}{X_\text{BBM}}
\newcommand{\xnn}{X_\text{NN}}
\newcommand{\datadim}{25}
\makeatletter
\begin{document}

\maketitle
\begin{abstract}
  Sufficient exploration is paramount for the success of a reinforcement
  learning agent. Yet, exploration is rarely assessed in an algorithm-independent way. We compare the behavior of three data-based,
  offline exploration metrics described in the literature on intuitive simple
  distributions and highlight problems to be aware of when using
  them. We propose a fourth metric, \emph{uniform relative entropy},
  and implement it using either a k-nearest-neighbor or a
  nearest-neighbor-ratio estimator, highlighting that the
  implementation choices have a profound impact on these measures.
\end{abstract}

\section{Introduction}

The problem of exploration vs.\ exploitation is one of the major
challenges in reinforcement learning: Should an agent choose
actions that maximize its reward (exploitation) or should it choose
actions that increase its knowledge while risking lower rewards
(exploration)?

Since the agent learns from the data it generates, its ability to
generate useful data limits its ability to achieve good
performance -- that is, if the agent explores insufficiently it will
not be able to learn a well-performing policy.

For distinct, given policies $\pi_i$ and $\pi_j$ on a given task the
exploitation is easy to judge: it is the expected return when the
actions are chosen according to $\pi_i$ or $\pi_j$ respectively. The
exploration is usually assessed only indirectly in terms of whether
the learned policy achieves good returns. An accurate quantification
of the exploration can indicate whether the agent is stuck in a local
optimum and whether further exploration could improve its performance.

Some algorithms (e.g.\ \cite{tangExplorationStudyCountBased2017,
  hongDiversitydrivenExplorationStrategy2018,
  burdaExplorationRandomNetwork2019,
  pongSkewFitStateCoveringSelfSupervised2020}) assess exploration
approximately by novelty or uncertainty measures. However, since those
algorithms directly try to optimize these measures and these measures
are algorithm-specific, it is difficult to use them to compare
exploration across algorithms.
While the ultimate goal of a reinforcement learning algorithm is to
find policies that achieve the highest reward (and eventually purely
exploit), achieving this is only possible if the algorithm explores
sufficiently and thereby collects the data to find admissible,
high-reward solutions. Thus measuring exploration could help in
understanding the reason for an algorithm's performance and can also
enable us to select, tune and debug them. This eventually will lead to
better-performing reinforcement learning methods.

During early training, with little information on the
maximally achievable reward or sparse reward signals, exploration
should be high to find these highly-rewarding regions in the state
space. Then the algorithm should start to explore around the regions
that have proven valuable and gradually shift towards exploitation,
while maintaining enough exploration for robust learning, as it moves
closer to the optimal solution.

In this work we propose to view the learning process of an RL agent as
a data-generating process and assess the achieved exploration offline
through the generated data $\mathcal{D}$, since this also allows
comparison across algorithms after the training has completed.

Without loss of generality, it is sufficient for the agent to reach
highly-rewarding regions in the state space. Note that if the reward
function depends on more than the current state, the state can be
augmented to include this information. We make no further assumptions
on where these highly-rewarding regions are.

We analyze three metrics from the literature (bin-count
$\xbin$~\cite{glassmanQuadraticRegulatorbasedHeuristic2010},
bounding-box-mean $\xbbm$~\cite{zhanTakingScenicRoute2019},
nuclear-norm $\xnn$~\cite{zhanTakingScenicRoute2019}) on four intuitive
state distributions. We highlight misleading results of these existing
metrics and propose our own metric $\xurel$ that outperforms these
existing metrics and provides a more accurate measure of exploration.

In our proposed metric we assume a uniform prior $\mathcal{U}$ on the
data generation for theoretically-maximal exploration, and judge the
achieved exploration of a collected dataset $\mathcal{D}$ by the
negative distance between this prior distribution and the generated
data distribution $\mathcal{Q}_\mathcal{D}$. This is our
\emph{uniform-relative-entropy} metric
\begin{equation}
\xurel(\mathcal{D}) =
  -\DKL\big(\mathcal{U}||\mathcal{Q}_\mathcal{D}\big) -
  \DKL\big(\mathcal{Q}_\mathcal{D}||\mathcal{U}\big).
\label{eq:xurel}
\end{equation}

\section{Method}

The learning process of an RL algorithm can be viewed as a data-generating
process that is run to produce a dataset of size $n$ sampled from an
unknown or implicit distribution $\mathcal{Q}_\mathcal{D}$:
\[\mathcal{D}^{(n)} \sim \mathcal{Q}_\mathcal{D}: \mathcal{D}^{(n)} = \{(s, a, r, s'), \ldots\}\]

\noindent where $(s, a, r, s')$ denotes a state, action, reward and
subsequent state tuple. Repeating the training process again draws a
new sample from this process. We assume that the state space of the
system forms a (hyper)box, i.e.\ the state-space
$S \in \mathbb{R} ^ d$ and is bounded by lower and upper limits:
$\forall s_i \in S: s_l \leq s_i \leq s_h $.

In such a case, the maximum-entropy distribution over the state space
is a uniform distribution
$\mathcal{U}$ \cite{udwadiaResultsMaximumEntropy1989}. We therefore
make the assumption that the highest exploration would be achieved
if the explored/generated data follows this uniform distribution and
measure this by the uniform relative entropy $\xurel$
\refeq{eq:xurel}.
In general, the relative-entropy (KL-divergence) term
$\DKL\big(P||Q\big)$ requires calculating a usually-intractable
integral, which has to be approximated by a sample estimate:
\begin{align}
  \DKL\big(P||Q\big) :=& \int p(s) \log \frac{p(s)}{q(s)} ds \\
\approx& \sum_{s \sim p} \log \frac{p(s)}{q(s)} \label{eq:approxkl}
\end{align}

Since the data distribution $\mathcal{Q}_\mathcal{D}$ in
\refeq{eq:xurel} is not available in closed form, the distance
$\DKL(\mathcal{U}\vert\vert\mathcal{Q}_\mathcal{D})$ has to be
estimated from the available sample $\mathcal{D}$. We assume that we
can sample from the prior and that log-likelihood values under the prior
are available for these samples. Thus, the KL-divergence can be
calculated using a density estimate of $\mathcal{D}$. This corresponds
to replacing $p(s)$ and $q(s)$ in \refeq{eq:approxkl} by density
estimates as necessary.

We looked at two ways of estimating the
divergence: \begin{enumerate*}[label=\textit{\alph*)}]
\item using an estimate for the density $\hat q(s)$ of
  $\mathcal{Q}_\mathcal{D}$ and
\item directly estimating the divergence using the ratio
  $\hat \rho(s)$, where $\rho(s)$ is the ratio between $p(s)$ and
  $q(s)$.
\end{enumerate*}
These will be described in the following two sections respectively.

We found that the estimators described in the next section
underestimate the divergence $\DKL\big(P||Q\big)$ when $P$ is
significantly more peaked than $Q$. Since we do not know apriori
whether $P$ or $Q$ will be more peaked, choosing either one could lead
to an inaccurate estimate in general. However, we found that using the
estimators for the symmetric KL-divergence compensates for this
problem and follows the actual KL-divergence, i.e.\ when $p(s)$ and
$q(s)$ are known, more closely.

\subsection{kNN estimator}\label{sec:knnestimator}

One possible estimator for $\hat q(s)$ is a k-Nearest-Neighbor (kNN) density
estimator \cite{bishopPatternRecognition2006}, where $V_d$ denotes the
unit volume of a $d$-dimensional sphere, $R_k(x)$ is the
Euclidean distance to the $k$-th neighbor of $x$, and $n$ is the total
number of samples in $\mathcal{D}$:
\begin{align}
  V_d = & \frac{ \pi^{d/2}}{\Gamma({\frac{d}{2} + 1})} \\
  \hat q_k(x) = & \frac{k}{n} \frac{1}{ V_d R_k(x)^d} =  \frac{k}{n V_d } \frac{1}{ R_k(x)^d}
\end{align}
where $\Gamma$ denotes the gamma-function.

\subsection{NNR estimator}

Noshad et al. \cite{noshadDirectEstimationInformation2017} proposed an
$f$-divergence estimator, the Nearest-Neighbor-Ratio (NNR) estimator,
based on the ratio of the nearest neighbors around a query point,
which we will use to estimate the KL-divergence.

They state that this estimator has a couple of beneficial properties.
Among these, the most important property for our metric is not being
affected by the over- / under-estimation artifacts of the kNN
estimator close to the boundaries of the state space.

For the general case of estimating $\DKL\big(P\vert\vert Q\big)$, we
take samples from $X \sim Q$ and $Y \sim P$. Let $\mathcal{R}_k(Y_i)$
denote the set of the $k$-nearest neighbors of $Y_i$ in the set
$Z:= X \cup Y$. $N_i$ is the number of points from
$X \cap \mathcal{R}_k(Y_i)$, $M_i$ is the number of points from
$Y \cap \mathcal{R}_k(Y_i)$, $M$ is the number of points in $Y$ and
$N$ is the number of points in $X$, $\eta = \frac{M}{N}$, and $C_L$ and
$C_U$ are the lower and upper limits of the densities $P$ and $Q$. Then,

\begin{align}
  \DKL(P\vert\vert Q) \approx & \hat D_g(X, Y) \\
  \hat D_g(X, Y) := & \max \left(
                      \frac{1}{M} \sum_{i=1}^M \tilde g \big(\frac{\eta N_i}{M_i + 1}, 0\big)
                      \right) \\
  \hat g(x) := & \max \big(g(x), g(C_L/C_U) \big) \\
  g(\rho) := & -\log(\rho)
\end{align}

\begin{figure*}[thb!]
\centering
\begin{subfigure}{1.0\textwidth}
  \centering
  \includegraphics[width=\textwidth]{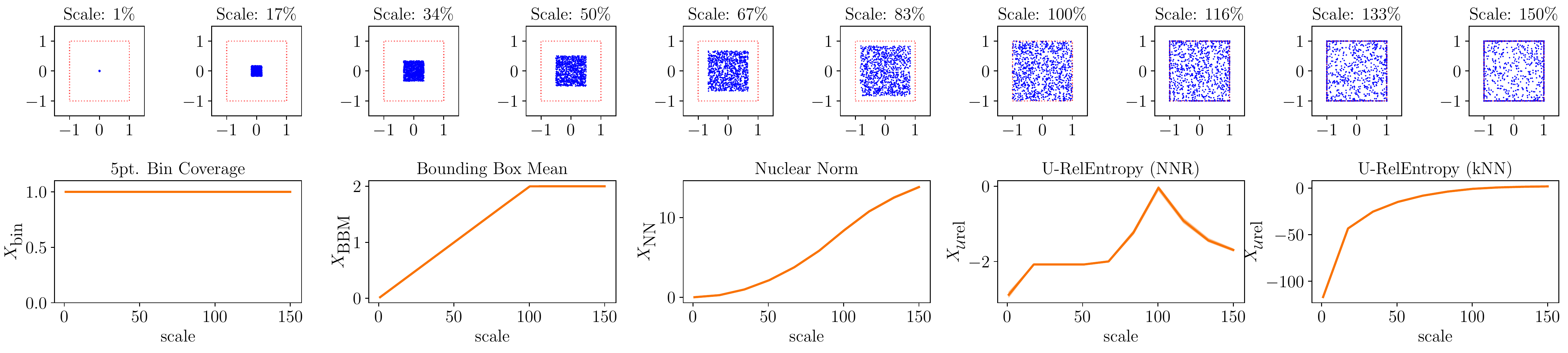}
\end{subfigure}
\begin{subfigure}{0.69\textwidth}
  \centering
  \subcaption{Growing Uniform distribution with values exceeding the
    state space clipped to the boundaries, thus an increasing number
    of points on the state-space boundaries for larger scales.
  }
  \label{fig:growing_uniform}
\end{subfigure}

\begin{subfigure}{1.0\textwidth}
  \centering
  \includegraphics[width=\textwidth]{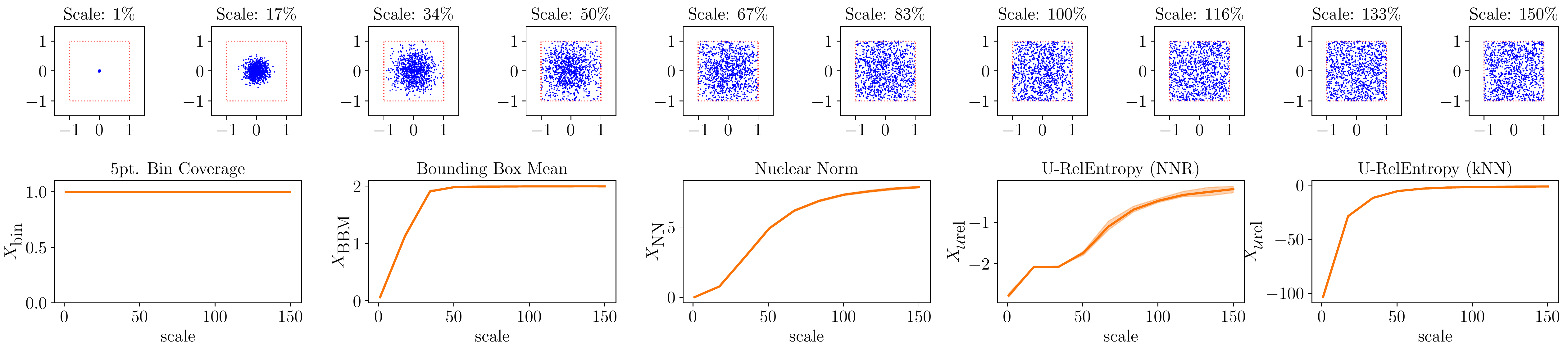}
  \subcaption{Growing Scale of Truncated Normal}
  \label{fig:truncated_normal}
  \vspace{1em}
\end{subfigure}

\begin{subfigure}{1.0\textwidth}
  \centering
  \includegraphics[width=\textwidth]{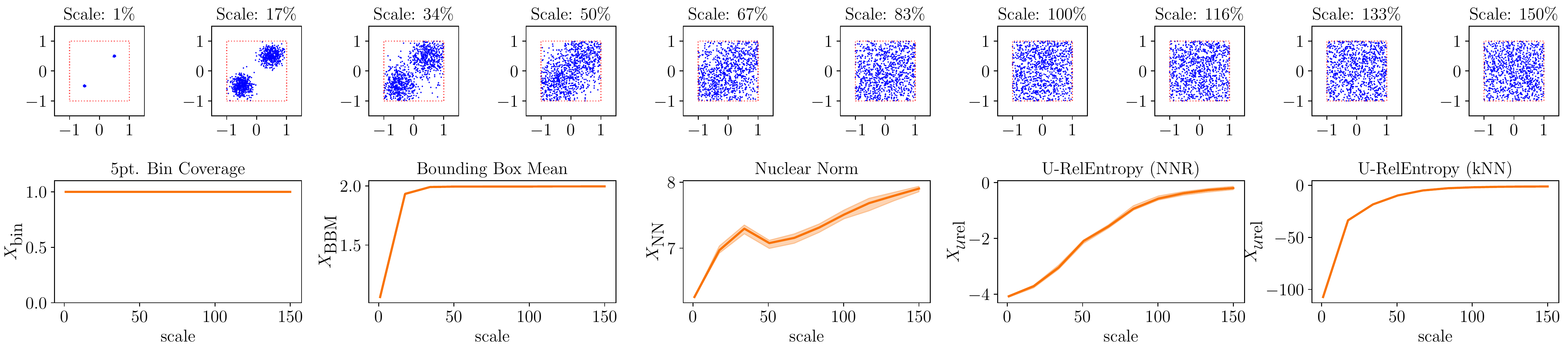}
  \subcaption{Growing scale of 2-Mixture of Truncated Normal}
  \label{fig:truncated_normal_scale}
  \vspace{1em}
\end{subfigure}

\begin{subfigure}{1.0\textwidth}
  \centering
  \includegraphics[width=\textwidth]{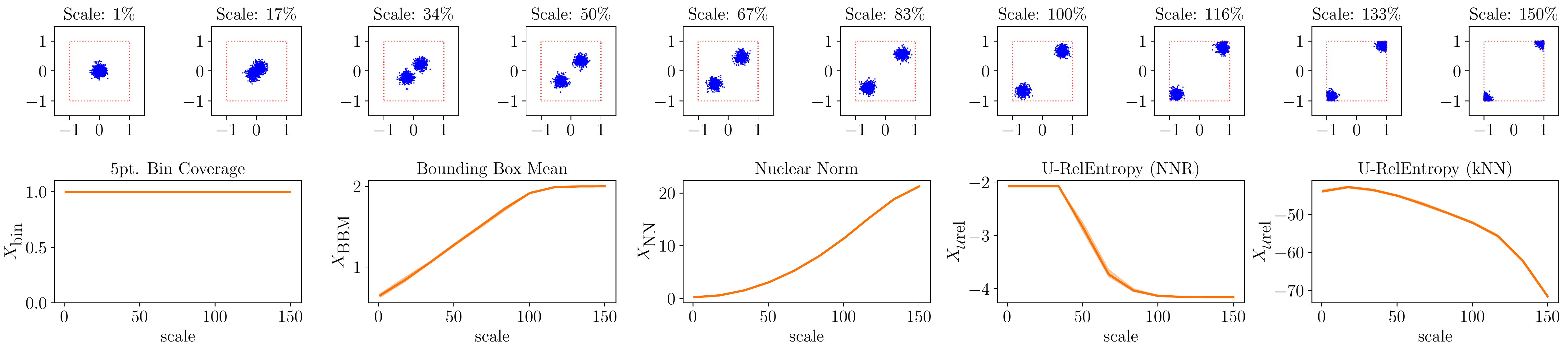}
  \subcaption{Growing Distance of Modes of 2-Mixture of Truncated Normal}
  \label{fig:truncated_normal_location}
  \vspace{1em}
\end{subfigure}

\caption{Comparison of exploration metrics on different data
  generating distributions dependent on one \emph{scale} parameter:
  $\datadim$ dimensional factorial distributions, similarly
  distributed along each dimension. The scatter plots depict first
  vs.\ second dimension (top a-d). Each comparison (bottom) shows the
  different exploration metrics $\xbin, \xbbm, \xnn$ and $\xurel$
  (ours) on the $y$ axis and the scale parameter is depicted on the
  $x$ axis.}
\label{fig:results}
\vspace{1em}
\end{figure*}
\section{Related Work}

Exploration can be measured by dividing the state space into equally
spaced bins and \emph{counting the percentage of non-empty bins}
\cite{glassmanQuadraticRegulatorbasedHeuristic2010}. This is related
to the directed count-based exploration metric in discrete
reinforcement-learning
settings$~$\cite{thrunEfficientExplorationReinforcement1992,bellemareUnifyingCountbasedExploration2016}
where each visited state acquires a bonus, for example
$\sum_i^n \sqrt{i}$ when it has been visited $n$ times. Thus, visiting
all states equally often in the limit achieves the highest exploration
bonus. This has also been extended to continuous domains by using
locality-sensitive hashing to discretize the state space
\cite{tangExplorationStudyCountBased2017}. We will compare our metric
$\xurel$ against this ratio of visited bins $\xbin$. We choose the
divisions such that we expect five data points in each bin on
average in the uniform case:
$\mathrm{divisions} = \left\lceil
  \left(\frac{N}{5}\right)^{\frac{1}{d}} \right\rceil$.  Since we
round up, the number of bins can exceed the total number of points $N$, to compensate we scale the
ratio by $\frac{1}{\min(N, \mathrm{divisions}})$ such that in the case
of more boxes than points the metric reaches its maximum when all
points lie in different boxes.

Zhan et al. \cite{zhanTakingScenicRoute2019} propose to measure the
sum of the side lengths of the hyperbox enclosing the collected data
$\mathcal{D}$ parallel to the state-space coordinate system, and they
denote this as the \emph{bounding-box-sum} metric. To reduce the
impact of the state-space dimensionality, we use the mean instead of
the sum of the bounding box side lengths and denote this as the
$\xbbm$ metric.

By its definition this metric is prone to over-estimating the covered
state space if the collected data points are not aligned with the
axes. So the authors \cite{zhanTakingScenicRoute2019} took this
problem into account and derived the \emph{nuclear-norm} metric $\xnn$
based on the sum of the eigenvalues (the trace) of the estimated
covariance $\mathbf {\hat C}$ of the data:
$ \xnn(\mathcal{D}) := \operatorname{\bf trace} \big( \mathbf{\hat
  C}(\mathcal{D}) \big) $.
\section{Evaluation}

To compare the different exploration metrics, we assumed a
$d=\datadim$-dimensional state space, generated data from four
different types of distributions, and compared the exploration metrics
on these data. The experiments were repeated $10$ times, and the mean
and min-max values are plotted.

These four cases are depicted in Figure \ref{fig:results}. While the
data are $d$-dimensional, they come from factorial distributions,
similarly distributed along each dimension. Thus, we can gain
intuition about the distribution from scatter plots of the first
vs.\ second dimension. This is depicted at the top of each of the four
parts. The bottom part of each comparison shows the different
exploration metrics, where the scale parameter is depicted on the $x$
axis and the exploration measure on the $y$ axis.

\paragraph{(a) Growing Uniform:} Figure \ref{fig:growing_uniform}
depicts data generated by a uniform distribution, centered around the
middle of the state space, with minimal and maximal values growing
relatively to the full state space according to the \emph{scale}
parameter from $1\%$ to $150\%$. Since in the latter case, many points
would lie outside the allowed state space; these values are clipped to
the state space boundaries. This loosely corresponds to an
undirectedly exploring agent that overshoots and hits the state space
limits, sliding along the state-space boundaries. Note how the
estimation (kNN vs. NNR) has a great impact on the $\xurel$ metric's
performance here: We would expect a maximum around a scale of $100\%$
and smaller values before and after (due to clipping). Here the
$\xurel$ (NNR) metric most closely follows this expectation. The
true diverge would follow a similar shape although with more extreme
values.  

\paragraph{(b) Truncated Normal:} Figure \ref{fig:truncated_normal}
depicts data of a truncated Gaussian distribution with the mean in the
center of the state-space box and the scale across all dimensions set
equal to the scale parameter. This example is inspired by random
exploration around a centered starting point. Since the distribution
is truncated, increasing the scale leads to more and more uniform
outcomes. The true divergence would converge to zero, which is
appropriately reflected in the metrics. The only difference to the
true divergence is that it again would exhibit more extreme low values and
rise more steeply. If an untruncated Gaussian distribution with
clipping was used, we would expect to see effects similar to Figure
\ref{fig:growing_uniform}.

\paragraph{(c) Bi-Modal Truncated Normal growing scale:} Figure~\ref{fig:truncated_normal_scale} shows a mixture of two truncated
  Gaussian distributions, spaced symmetrically around the center of
  the state space, with growing standard deviations set equal to the
  scale parameter. This can be compared to cases where the search
  process is initiated from certain fixed starting
  positions. Increasing the scale again leads to more and more uniform
  behavior. Note that the bounding-box mean $X_\text{BBM}$ rapidly
  approaches the maximum value and provides no further information. A
  further interesting artifact is visible in the nuclear norm
  $X_\text{NN}$, where the measured exploration drops without an
  intuitive explanation.
 
\paragraph{(d) Bi-Modal Truncated Normal moving locations:} Figure
  \ref{fig:truncated_normal_location} shows a mixture of two truncated
  Gaussian distributions, with equal standard deviations but located
  further and further apart (depending on the scale parameter). In
  this case, the state space coverage should increase until both
  distributions are sufficiently apart, should then stay the same, and
  begin to drop as the proximity to the state space limits the points
  to an ever smaller volume. While somewhat contrived, it highlights
  difficulties in the exploration metrics. Both the bounding-box mean
  $X_\text{BBM}$ and the nuclear norm $X_\text{NN}$ completely fail to
  account for vastly unexplored areas between the extreme points.
Since the $\xurel$ NNR metric is clipped (by definition of NNR) the
  metric reaches its limits when the density ratios become extreme,
  which presumably happens for very small and large scale parameters
  in this setting. Here the $X_\text{Urel}$ kNN appears to outperform
  NNR even though it also suffers from under-estimation of the
  divergence for points close to the boundaries.
\section{Conclusion}

We compared four exploration metrics on generated data and highlighted
shortcomings and caveats of these metrics: \begin{enumerate*}[label=\textit{(\roman*)}]
\item Our $\xurel$-NNR metric correctly shows a decrease in
  exploration when points are clipped to the state space boundaries,
  whereas the other metrics do not detect this problem,
\item $\xurel$ converges to zero as the state space coverage becomes
  more uniform as shown for the growing truncated normal
  distributions,
\item while $\xbin$ is shown not to be useful in high-dimensions,
  $\xurel$ is.
\item $\xbbm$ rapidly reaches the maximum and does not provide any
  useful information and $\xnn$ over-estimates the importance of the
  most spread out points.
  \end{enumerate*}
We propose our metric using the uniform distribution $\mathcal{U}$ as
the most general case. However, in some cases more information about
the task may be available. For example, when a rough estimate of the
goal location is already known, then it is more reasonable to explore
around that location, rather than uniformly over the whole state
space. In such a case a more appropriate prior should be selected.

These metrics could also be used for goal babbling: consider a
redundant robot arm, where the relevant aspect would be the reached
configurations of the end-effector rather than reaching all possible
joint-space configurations. In this case, transforming the collected
data into the goal space, i.e.\ the end-effector configurations, and
estimating the metrics in that space would provide an effective
measurement of exploration.

We showed that the choice of metric, as well as the implementation
details (such as the kNN vs.\ NNR estimator) greatly influence the
behavior of the metric, and that, since approximations are necessary,
an intuitive understanding of the metrics is beneficial.

Our metric is designed to be used offline, to provide information
about the exploration performance across algorithms. This allows us to
approximate the relative entropy more accurately than would be
possible inside the training loop.

In future work we will use this metric to analyze the learning
progress of various deep reinforcement learning algorithms. We will
also investigate whether this can help detect why learning sometimes
fails.

\section*{Acknowledgements}

The research leading to these results has received funding from the
European Union’s Horizon 2020 research and innovation programme under
grant agreement no. 731761, IMAGINE.

\printbibliography

\end{document}